\documentclass{article}

\usepackage[final]{neurips_2021_imagenet}

\usepackage[utf8]{inputenc} 
\usepackage[T1]{fontenc}    
\usepackage{hyperref}       
\usepackage{url}            
\usepackage{booktabs}       
\usepackage{amsfonts}       
\usepackage{nicefrac}       
\usepackage{microtype}      
\usepackage{xcolor}         
\usepackage{threeparttable}
\usepackage{natbib}
\usepackage[pdftex]{graphicx}

\title{NAM: Normalization-based Attention Module}

\author{%
  Yichao Liu
    \\
  Helmholtz-Zentrum Dresden-Rossendorf\\
  Dresden, Germany\\
  \texttt{y.liu@hzdr.de}  \\
  \And
  Zongru Shao \\
  Helmholtz-Zentrum Dresden-Rossendorf \\
  Dresden, Germany \\
  Center for Advanced Systems Understanding \\
  Görlitz, Germany \\
  \texttt{z.shao@hzdr.de} \\
  \AND
  Yueyang Teng \\
  College of Medicine and Biological Information Engineering\\
  Northeastern University \\
  110004 Shenyang, China \\
  \texttt{tengyy@bmie.neu.edu.cn} \\
  \And
  Nico Hoffmann \\
  Helmholtz-Zentrum Dresden-Rossendorf \\
  Dresden, Germany \\
  \texttt{n.hoffmann@hzdr.de} \\
}

\begin{document}

\maketitle

\begin{abstract}

Recognizing less salient features is the key for model compression. However, it has not been investigated in the revolutionary attention mechanisms. In this work, we propose a novel normalization-based attention module (NAM), which suppresses less salient weights. It applies a weight sparsity penalty to the attention modules, thus, making them more computational efficient while retaining similar performance. A comparison with three other attention mechanisms on both Resnet and Mobilenet indicates that our method results in higher accuracy. Code for this paper can be publicly accessed at \url{https://github.com/Christian-lyc/NAM}. 

\end{abstract}

\section{Introduction}

Attention mechanisms have been one of the heated research interests in recent years (\citet{wang2017residual, hu2018squeeze, park2018bam, woo2018cbam, gao2019global}). It assists deep neural networks to suppress less salient pixels or channels. Many of the prior studies focus on capturing salient features with attention operations (\citet{zhang2020relation,misra2021rotate}). Those methods successfully exploit the mutual information from different dimensions of features. However, they lack consideration on the contributing factors of weights, which is capable of further suppressing the insignificant channels or pixels. Inspired by \citet{liu2017learning}, we aim to utilize the contributing factors of weights for the improvement of attention mechanisms. We use a scaling factor of batch normalization which uses the standard deviation to represent the importance of weights. This can avoid adding fully-connected and convolutional layers, which is used in the SE, BAM and CBAM. Thus, we propose an efficient attention mechanism -- Normalization-based Attention Module (NAM).

\section{Related work}

Many prior works attempt to improve the performance of neural networks by suppressing insignificant weights. Squeeze-and-Excitation Networks (SENet) (\citet{hu2018squeeze}) integrate the spatial information into channel-wise feature responses and compute the corresponding attention with two multi-layer-perceptron (MLP) layers. Later, Bottleneck Attention Module (BAM) (\citet{park2018bam}) builds separated spatial and channel submodules in parallel and they can be embedded into each bottleneck block. Convolutional Block Attention Module (CBAM) (\citet{woo2018cbam}) provides a solution that embeds the channel and spatial attention submodules sequentially. To avoid the ignorance of cross-dimension interactions, Triplet Attention Module (TAM) (\citet{misra2021rotate}) takes account of dimension correlations by rotating the feature maps. However, these works neglect information from the tuned weights from training. Therefore, we aim to highlight salient features by utilizing the variance measurement of the trained model weights.

\section{Methodology}
\begin{figure}[ht!]
  \centering
  \includegraphics[width=0.8\linewidth]{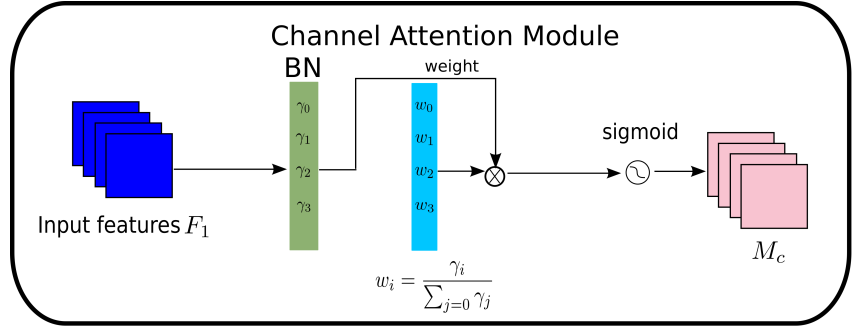}
  \caption{Channel attention mechanism}
  \label{fig:2}
\end{figure}

\begin{figure}[ht!]
  \centering
  \includegraphics[width=0.8\linewidth]{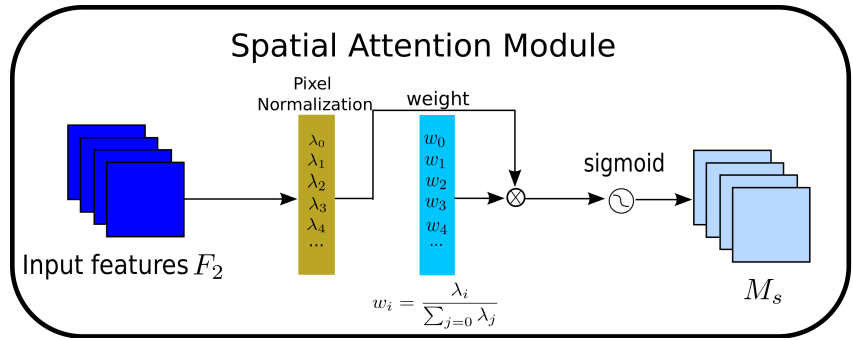}
  \caption{Spatial attention mechanism}
  \label{fig:3}
\end{figure}

We propose NAM as an efficient and lightweight attention mechanism. We \textbf{adopt the module integration from CBAM} (\citet{woo2018cbam}) and redesign the channel and spatial attention submodules. Then, a NAM module is embedded at the end of each network block. For residual networks, it is embedded at the end of the residual structures. 
For the \textbf{channel attention} submodule, we use a scaling factor from batch normalization (BN) (\citet{ioffe2015batch}), as shown in Equation (\ref{eq:bn}). The scaling factor measures the variance of channels and indicates their importance. 

\begin{equation}
    \label{eq:bn}
    B_{out}=BN(B_{in})=\gamma \frac{B_{in}-\mu_{\mathcal{B}}}{\sqrt{\sigma ^{2}_{\mathcal{B}}+\epsilon} }+\beta
\end{equation}

where $\mu_\mathcal{B}$ and $\sigma_\mathcal{B}$ are the mean and standard deviation of mini batch $\mathcal{B}$, respectively; $\gamma$ and $\beta$ are trainable affine transformation parameters (scale and shift) (\citet{ioffe2015batch}). The channel attention submodule is shown in Figure \ref{fig:2} and Equation (\ref{eq:1}), where $\mathbf{M}_{c}$ represents the output features. $\gamma$ is the scaling factor for each channel, and the weights are obtained as $W_{\gamma}=\gamma_{i}/\sum_{j=0} \gamma_{j}$. 
We also apply a scaling factor of BN to the spatial dimension to measure the importance of pixels. We name it pixel normalization.
The corresponding \textbf{spatial attention} submodule is shown in Figure \ref{fig:3} and Equation (\ref{eq:2}), where the output is denoted as $\mathbf{M}_{s}$. $\lambda$ is the scaling factor, and the weights are $W_{\lambda}=\lambda_{i}/\sum_{j=0} \lambda_{j}$.

To suppress the less salient weights, we add a regularization term into the loss function, as shown in Equation (\ref{eq:3}) (\citet{liu2017learning}), where $x$ denotes the input; $y$ is the output; $W$ represents network weights; $l(\cdot)$ is the loss function; $g(\cdot)$ is the $l_{1}$ norm penalty function; $p$ is the penalty that balances $g(\gamma)$ and $g(\lambda)$.
 
 \begin{equation}
     \mathbf{M}_{c}=sigmoid(W_{\gamma}(BN(\mathbf{F}_{1})))
     \label{eq:1}
 \end{equation}
  \begin{equation}
     \mathbf{M}_{s}=sigmoid(W_{\lambda}(BN_{s}(\mathbf{F}_{2})))
     \label{eq:2}
 \end{equation}
  \begin{equation}
     Loss=\sum _{(x,y)} l(f(x,W),y)+ p\sum g(\gamma)+p\sum g(\lambda)
     \label{eq:3}
 \end{equation}

 \section{Experiment}
 \label{Experiment}
In this section, we compare the performance of NAM with SE, BAM, CBAM, and TAM for ResNet and MobileNet. We evaluate every method with four Nvidia Tesla V100 GPUs on a cluster. We first run ResNet50 on CIFAR-100 (\citet{krizhevsky2009learning}) and use the same preprocess and training configurations as CBAM (\citet{woo2018cbam}), with $p$ as 0.0001. The comparison in Table \ref{table:1} indicates that NAM with channel or spatial attention alone outperforms the other four attention mechanisms. 
We then run MobileNet on ImageNet (\citet{deng2009imagenet}) as it is one of the standard datasets for image classification benchmarks. We set $p$ as 0.001 and the rest of the configurations the same as CBAM. The comparison in Table \ref{table:2} shows that NAM with channel and spatial attention combined outperforms the other three with similar computation complexity.

\begin{table}[ht!]
  \caption{Classification results on CIFAR-100}
  \label{table:1}
  \centering
  \begin{tabular}{lcccc}
    \toprule
    Architecture     & Parameters     & FLOPs  & Top-1 Error ($\%$) & Top-5 Error ($\%$)\\ %
    \midrule

    ResNet 50 & 23.71M  & 1.30G &  22.74 & 6.37  \\ %
    ResNet 50 + SE  & 26.22M & 1.31G  & 20.29 & 5.18   \\ %
    ResNet 50 + BAM & 24.06M & 1.33G  & 19.97 &  5.03   \\ %
    ResNet 50 + CBAM  & 26.24M & 1.31G & 19.44 & 4.66    \\ %
    ResNet 50 + TAM & 23.71M & 1.33G & 20.15 & 5.13 \\
    ResNet 50 + NAM(ch*)  & 23.74M & 1.31G & 19.09 & 4.5   \\ %
    ResNet 50 + NAM(sp*)  & 23.71M & 1.31G & 19.38 & 4.72  \\ %
    \bottomrule
  \end{tabular}
  \begin{tablenotes}
     \item $\star$ ch stands for channel attention only; sp indicates spatial attention only.
\end{tablenotes}
\end{table}

\begin{table}[ht!]
  \caption{Classification results on ImageNet}
  \label{table:2}
  \centering
  \begin{tabular}{lcccc}
    \toprule
    Architecture    & Parameters     & FLOPs  & Top-1 Error ($\%$) & Top-5 Error ($\%$) \\ %
    \midrule

    MobileNet V2   & 3.51M & 0.31G   & 30.52 & 11.20 \\ %
    MobileNet V2 + SE & 3.53M & 0.32G & 29.77 &  10.65   \\ %
    MobileNet V2 + BAM  & 3.54M & 0.32G  & 29.91 & 10.80   \\ %
    MobileNet V2 + CBAM & 3.54M & 0.32G & 29.74 & 10.66    \\ %
    MobileNet V2 + NAM & 3.51M & 0.32G  & 29.34 & 10.18   \\ %
    \bottomrule
  \end{tabular}
\end{table}

\section{Conclusion}
\label{Conclusion}

We proposed a NAM module that is more efficient by suppressing the less salient features. Our experiments indicate that NAM provides efficiency gain on both ResNet and MobileNet. We are conducting a detailed analysis of the performance of NAM regarding its integration variations and hyper-parameter tuning. We also plan to optimize NAM with different model compression techniques to promote its efficiency. In the future, we will investigate its effects on other deep learning architectures and applications.

\bibliographystyle{unsrtnat}
\bibliography{citation}

\begin{thebibliography}{11}
\providecommand{\natexlab}[1]{#1}
\providecommand{\url}[1]{\texttt{#1}}
\expandafter\ifx\csname urlstyle\endcsname\relax
  \providecommand{\doi}[1]{doi: #1}\else
  \providecommand{\doi}{doi: \begingroup \urlstyle{rm}\Url}\fi

\bibitem[Wang et~al.(2017)Wang, Jiang, Qian, Yang, Li, Zhang, Wang, and
  Tang]{wang2017residual}
Fei Wang, Mengqing Jiang, Chen Qian, Shuo Yang, Cheng Li, Honggang Zhang,
  Xiaogang Wang, and Xiaoou Tang.
\newblock Residual attention network for image classification.
\newblock In \emph{Proceedings of the IEEE conference on computer vision and
  pattern recognition}, pages 3156--3164, 2017.

\bibitem[Hu et~al.(2018)Hu, Shen, and Sun]{hu2018squeeze}
Jie Hu, Li~Shen, and Gang Sun.
\newblock Squeeze-and-excitation networks.
\newblock In \emph{Proceedings of the IEEE conference on computer vision and
  pattern recognition}, pages 7132--7141, 2018.

\bibitem[Park et~al.(2018)Park, Woo, Lee, and Kweon]{park2018bam}
Jongchan Park, Sanghyun Woo, Joon-Young Lee, and In~So Kweon.
\newblock Bam: Bottleneck attention module.
\newblock \emph{arXiv preprint arXiv:1807.06514}, 2018.

\bibitem[Woo et~al.(2018)Woo, Park, Lee, and Kweon]{woo2018cbam}
Sanghyun Woo, Jongchan Park, Joon-Young Lee, and In~So Kweon.
\newblock Cbam: Convolutional block attention module.
\newblock In \emph{Proceedings of the European conference on computer vision
  (ECCV)}, pages 3--19, 2018.

\bibitem[Gao et~al.(2019)Gao, Xie, Wang, and Li]{gao2019global}
Zilin Gao, Jiangtao Xie, Qilong Wang, and Peihua Li.
\newblock Global second-order pooling convolutional networks.
\newblock In \emph{Proceedings of the IEEE/CVF Conference on Computer Vision
  and Pattern Recognition}, pages 3024--3033, 2019.

\bibitem[Zhang et~al.(2020)Zhang, Lan, Zeng, Jin, and Chen]{zhang2020relation}
Zhizheng Zhang, Cuiling Lan, Wenjun Zeng, Xin Jin, and Zhibo Chen.
\newblock Relation-aware global attention for person re-identification.
\newblock In \emph{Proceedings of the ieee/cvf conference on computer vision
  and pattern recognition}, pages 3186--3195, 2020.

\bibitem[Misra et~al.(2021)Misra, Nalamada, Arasanipalai, and
  Hou]{misra2021rotate}
Diganta Misra, Trikay Nalamada, Ajay~Uppili Arasanipalai, and Qibin Hou.
\newblock Rotate to attend: Convolutional triplet attention module.
\newblock In \emph{Proceedings of the IEEE/CVF Winter Conference on
  Applications of Computer Vision}, pages 3139--3148, 2021.

\bibitem[Liu et~al.(2017)Liu, Li, Shen, Huang, Yan, and Zhang]{liu2017learning}
Zhuang Liu, Jianguo Li, Zhiqiang Shen, Gao Huang, Shoumeng Yan, and Changshui
  Zhang.
\newblock Learning efficient convolutional networks through network slimming.
\newblock In \emph{Proceedings of the IEEE international conference on computer
  vision}, pages 2736--2744, 2017.

\bibitem[Ioffe and Szegedy(2015)]{ioffe2015batch}
Sergey Ioffe and Christian Szegedy.
\newblock Batch normalization: Accelerating deep network training by reducing
  internal covariate shift.
\newblock In \emph{International conference on machine learning}, pages
  448--456. PMLR, 2015.

\bibitem[Krizhevsky et~al.(2009)Krizhevsky, Hinton,
  et~al.]{krizhevsky2009learning}
Alex Krizhevsky, Geoffrey Hinton, et~al.
\newblock Learning multiple layers of features from tiny images.
\newblock 2009.

\bibitem[Deng et~al.(2009)Deng, Dong, Socher, Li, Li, and
  Fei-Fei]{deng2009imagenet}
Jia Deng, Wei Dong, Richard Socher, Li-Jia Li, Kai Li, and Li~Fei-Fei.
\newblock Imagenet: A large-scale hierarchical image database.
\newblock In \emph{2009 IEEE conference on computer vision and pattern
  recognition}, pages 248--255. Ieee, 2009.

\end{thebibliography}

\appendix
\section{Appendix}
\subsection{Comparison of CBAM and NAM regarding the number of parameters}

\begin{table}[ht!]
  \caption{The comparison of channel attention for CBAM and NAM on ResNet50}
  \label{table:4}
  \centering
  \begin{tabular}{lcc}
    \toprule
  Parameters & CBAM  & NAM  \\ %
    \midrule
    Block1 & 512*4*512*4/16*2 ($C*R*C*R/r*2$)  &  512*4 ($C*R$)  \\ %
    Block2  & 256*4*256*4/16*2 &  256*4 \\ %
    Block3  & 128*4*128*4/16*2 &  128*4 \\ %
    Block4  & 64*4*64*4/16*2 &  64*4 \\ %
    Overhead & 696320  &  3840 \\
    \bottomrule
  \end{tabular}
\end{table}

We show a comparison of the number of parameters in CBAM and NAM in Table \ref{table:3} and \ref{table:4}.  They empirically verify the parameter reduction of NAM. In the channel attention module, $C$ represents the number of the input channels of each block. $R$ represents the expanding ratio of each block. $r$ represents the reduction ratio utilized in the MLP to compute the channel attention, which is set to 16 in CBAM. The kernel size is denoted as $k$, which is 7. In NAM, $H$ and $W$ represent the height and width of the input images respectively. From Table \ref{table:3} and \ref{table:4}, we observe a significant parameter reduction in the channel attention module and an insignificant increase of parameters in the spatial attention module of NAM against CBAM. As a result, NAM has fewer parameters than CBAM.

\begin{table}[ht!]
  \caption{The comparison of spatial attention for CBAM and NAM on ResNet50}
  \label{table:3}
  \centering
  \begin{tabular}{lcc}
    \toprule
  Parameters & CBAM  & NAM  \\ %
    \midrule

    Block1 & 2*1*7*7 ($2*1*k^2$)  &  32*32 ($H*W$)  \\ %
    Block2  & 2*1*7*7 &  16*16 \\ %
    Block3  & 2*1*7*7 &  8*8 \\ %
    Block4  & 2*1*7*7 &  4*4 \\ %
    Overhead & 392    &  1360 \\
    \bottomrule
  \end{tabular}
\end{table}

\end{document}